\documentclass{article}
\usepackage{arxiv}
\usepackage{graphicx}
\usepackage[numbers]{natbib}
\usepackage{doi}
\usepackage[utf8]{inputenc} 
\usepackage[T1]{fontenc}    
\usepackage{hyperref}       
\usepackage{url}            
\usepackage{booktabs}       
\usepackage{amsfonts}       
\usepackage{nicefrac}       
\usepackage{microtype}      
\usepackage[table]{xcolor}         
\usepackage{amsmath}
\usepackage{amssymb}
\usepackage{array}
\usepackage{pifont}
\usepackage{threeparttable}
\usepackage{graphicx} 
\usepackage{array} 
\usepackage{makecell}
\usepackage{algpseudocode} 
\usepackage{multirow}
\usepackage{listings}
\usepackage{enumitem}
\usepackage{subfig}
\usepackage{hyperref}
\usepackage{todonotes}
\usepackage{float}
\usepackage[ruled,vlined]{algorithm2e}

\usepackage{authblk}

\newcommand{\pangu}{Pangu}
\newcommand{\modelname}{\pangu~Ultra}

\title{\modelname: Pushing the Limits of Dense Large Language Models on Ascend NPUs}

\author{\pangu~Team, Huawei\\
PanguTech@huawei.com}

\renewcommand{\headeright}{Technical Report}

\begin{document}

\maketitle

\thispagestyle{fancy}

\begin{abstract}

We present \modelname, a Large Language Model (LLM) with 135 billion parameters and dense Transformer modules trained on Ascend Neural Processing Units (NPUs). Although the field of LLM has been witnessing unprecedented advances in pushing the scale and capability of LLM in recent years, training such a large-scale model still involves significant optimization and system challenges. To stabilize the training process, we propose depth-scaled sandwich normalization, which effectively eliminates loss spikes during the training process of deep models. We pre-train our model on 13.2 trillion diverse and high-quality tokens and further enhance its reasoning capabilities during post-training. To perform such large-scale training efficiently, we utilize 8,192 Ascend NPUs with a series of system optimizations. Evaluations on multiple diverse benchmarks indicate that \modelname~significantly advances the state-of-the-art capabilities of dense LLMs such as Llama 405B and Mistral Large
2, and even achieves competitive results with DeepSeek-R1, whose sparse model structure contains much more parameters. Our exploration demonstrates that Ascend NPUs are capable of efficiently and effectively training dense models with more than 100 billion parameters. 
Our model and system will be available for our commercial customers.
\end{abstract}

\section{Introduction}

Large Language Models (LLMs) have transformed the landscape and our understanding of Artificial Intelligence. Their remarkable capabilities are enabling more and more AI applications, bringing numerous commercial opportunities. Unsurprisingly, teams are racing to push the scaling law to create models with more and more parameters. Although the Transformer~\cite{vaswani2017attention} structure is a popular choice for large models, it is still debatable whether the models should be sparse or dense. With more than 100 billion parameters, sparse architectures powered by Mixture of Experts (MoE), such as DeepSeek \cite{liu2024deepseek,deepseekv3}, have demonstrated surreal human-like language and thinking abilities~\cite{jones2025large}, which makes sparse models a pupular choice when pushing the limit of LLMs.

At the same time, dense models, such as the Qwen~\cite{bai2023qwen,yang2024qwen2}, Llama~\cite{grattafiori2024llama}, and Gemma \cite{team2024gemma} series, are currently popular among models with fewer than 100 billion parameters thanks to their strong performance in specific skills and ease of deployment. The parameters in dense models are usually easier to optimize, while the dynamic components in sparse models usually need to turn to additional heuristics for stable training.
In addition, the dense model structures at inference time make it easier to optimize system performance due to deterministic parameter usage. In this study, we aim to further explore the potential of dense models at large scales and show the performance of dense models can be on par with state-of-the-art MoE models on diverse tasks.

The numbers of model parameters and layers are two crucial dimensions to release the full potential of dense models. While model parameter count is critical for model performance and plays a central role in scaling laws~\cite{kaplan2020scaling}, recent studies~\cite{ye2024physics,merrill2025littledepth} suggest that model depth has a significant impact on reasoning capabilities. However, our exploration in those two aspects poses significant challenges in exploring the limits of those two aspects. Deeper models usually introduce unstable training, manifested as spikes in training loss curves. Experimental observations suggest that those spikes can knock our model out of the ideal parameter landscape and cause irreparable damage to the training process. Meanwhile, training hundreds of billions of parameters in dense models requires orchestrating thousands of AI processors, which poses significant system efficiency challenges.

For our exploration, we introduce \modelname, a dense Transformer architecture with 135 billion parameters and 94 layers. The model setup is at the forefront scale of the top performing dense models~\cite{bai2023qwen,yang2024qwen2,grattafiori2024llama,team2024gemma}. Regarding challenges of training deep models, we hypothesize that the loss spikes are due to gradient fluctuations, which in turn hinder convergence rates and may lead to training divergence. Therefore, we propose two techniques, the depth-scaled sandwich norm and tiny initialization, both of which are designed to maintain stable gradient norms. Specifically, we first replace pre-layer norm \cite{DBLP:conf/emnlp/LiuLGCH20} with the sandwich norm \cite{NEURIPS2021_CogView} and scaled initialization values in the post-layer normalization based on the model's depth. This depth-based adjustment helps control the range of gradient fluctuations effectively. In addition, we scale the standard deviation of weight initialization according to the model's width and depth, leading to tiny initialization. These two techniques lead to more stable gradients throughout the training process, eliminating loss spikes during the training of \modelname, and improving overall model performance.

In practice, we pre-train \modelname~on 13.2 trillion tokens of our built corpus. In the pre-training stage, we use three phrases of data corpus each with a distinct data recipe. The design principles behind three phrases are first to help the model develop knowledge and linguistics, and then to directly equip it with reasoning ability, and finally to boost it on actively learning to reason. The model context window is gradually extended from 4K to 128K. In the post-training stage, we begin with applying efficient supervised fine-tuning (SFT) for a cold start, utilizing a carefully curated set of instruction data. Following this, \modelname~undergoes further optimization through Reinforcement Learning (RL). The overall training of \modelname~is stable in this process.

To handle large-scale model training of more than 100 billion parameters, we utilize a large-scale computing cluster consisting of 8,192 Ascend NPUs and employ a series of system optimization to improve the system efficiency.
The primary challenge is minimizing pipeline bubbles~\cite{huang2019gpipe} at large scales, which arise due to batch size constraints~\cite{megascale}. 
We take advantage of the typical 4 types of parallelism on our Ascend cluster, that is, Data Parallelism (DP), Tensor Parallelism (TP)~\cite{megatron-lm}, Sequence Parallelsim \cite{megatron-sp} and Pipeline Parallelism (PP)~\cite{gpipe,pipedream}. 
As the training cluster scales up, the mini-batch size allocated to each DP decreases, leading to an increased pipeline bubble ratio. To mitigate this issue, we employ additional virtual pipeline (VPP) scheduling \cite{interleaved-pp} with fine-grained tuning to ensure load balancing and reduce the PP bubble ratio from 30.45\% to 6.8\%.
The second challenge is to achieve high training efficiency for long sequences. 
Both attention mask generation and self-attention computation are time- and memory-intensive, particularly for long contexts.
We utilize a NPU Fusion Attention (NFA) operator \cite{fa-github, flash_attention, flash_attention2} tailored for the Ascend NPUs, which supports reset attention mask scenarios and eliminates the need to construct the attention mask before calling the NFA, thus improving computational efficiency and reducing memory cost.
Under the implementation of several fine-grained system optimization, we achieve a Model FLOPs Utilization (MFU) \cite{palm} of over 50\% when training \modelname~on 8,192 Ascend NPUs.

On public evaluation benchmarks, \modelname~outperforms existing dense LLMs including Llama 405B and Mistral Large 2 123B on almost all major language tasks, and achieves competitive results with sparse models consisting of more than 500 billion parameters. These results indicate the potential of dense model capabilities is still promising to explore. \modelname~also demonstrates that the Ascend NPUs are suitable for exploring the full capabilities of large-scale dense language models.

\section{Model Architecture}

The basic architecture of \modelname~is similar to Llama 3~\cite{grattafiori2024llama}. It has 135 billion parameters with a hidden dimension of 12,288, a SwiGLU~\cite{shazeer2020glu} feed-forward network (FFN) intermediate size of 28,672, and 94 layers. The attention blocks in \modelname~leverage Group Query Attention (GQA) to reduce KV-cache size by incorporating 96 query heads and 8 KV heads.

There are two crucial differences to address the fundamental challenges of training stability and convergence in large dense LLMs. We propose Depth-Scaled Sandwich-Norm to replace the layer normalization and TinyInit for parameter initialization. By integrating these techniques, \modelname~achieves substantial improvements over previous dense models.

\subsection{Depth-Scaled Sandwich-Norm}

Large-scale dense models typically adopt deeper architectures \cite{dubey2024llama}, although MoE models usually scale in width \cite{deepseekv3}. However, increased depth introduces greater challenges in maintaining training stability. 
Given the prohibitive cost of pre-training, stable training of large dense LLMs becomes paramount.
Pre-Layer Normalization (Pre-LN) has been found to make back-propagation more efficient for deep Transformers \cite{wang2019learning-deep}, leading to its widespread adoption in Transformer-based large language model (LLM) architectures \cite{dubey2024llama,bai2023qwen,deepseekv3}.

However, in models employing the pre-LN structure, the fluctuating output scale of each sub-layer can easily lead to training instability \cite{takase2024spikemore}.
To address this issue, sandwich-norm \cite{NEURIPS2021_CogView} applies an layer normalization to each sub-layer's output prior to the residual connection. While the sandwich-norm maintains the scale stability of individual sub-layer outputs, the progressive accumulation of output norms via residual connections across multiple layers may nevertheless lead to training instability.

\begin{figure}[htbp]
    \centering
    \includegraphics[width=0.8\textwidth]{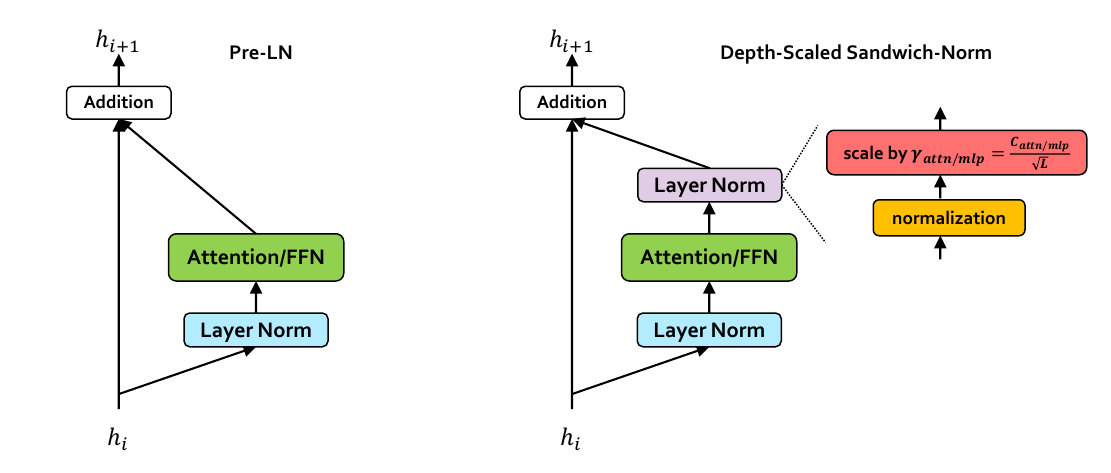}
    \caption{Structure comparison between Pre-Layer Norm (Pre-LN) and Depth-Scaled Sandwich-Norm (DSSN). DSSN applies normalization layers to both before and after the attention and FFN block, while Pre-LN only utilizes one normalization layer. DSSN also employs a depth-scaled initialization schema, which is not in the original sandwich norm.
    }
    \label{fig:sandwich_arch}
\end{figure}

To mitigate this, we present the depth-scaled sandwich norm, which integrates the sandwich norm with a depth-scaled initialization scheme. The layer normalization regulates layer-wise output magnitudes through trainable gamma parameters, which are initialized with values scaled proportionally to the inverse of network depth. Figure ~\ref{fig:sandwich_arch} illustrates the differences between the depth-scaled sandwich-norm and pre-norm architectures. The formula of depth-scaled sandwich-norm is

\begin{equation} \label{eq:sandwich}
\begin{aligned}
    \mathbf{h} &\leftarrow \mathbf{h} + {\text{Norm}}({\gamma_{\text{attn}}}, \text{ATTN}(\text{Norm}(\mathbf{h}))), \quad  \gamma_{\text{attn}} = \frac{c_{\text{attn}}}{\sqrt{L}},\\ 
    \mathbf{h} &\leftarrow \mathbf{h} + {\text{Norm}}({\gamma_{\text{mlp}}}, \text{MLP}(\text{Norm}(\mathbf{h}))), \quad \gamma_{\text{mlp}} = \frac{c_{\text{mlp}}}{\sqrt{L}}, \\
\end{aligned}
\end{equation}

where $L$ is the number of layers, $c_{\text{attn}}$ and $c_{\text{mlp}}$ are set as the initial output standard deviations of the attention layer and feed-forward network (FFN) layer, respectively. For \modelname, we set $c_{\text{attn}}$ to 0.283 and $c_{\text{mlp}} $ to 0.432 .

\subsection{Model Initialization} \label{sec:tinyinit}

Existing works \cite{nguyen2019transformers} observe that model initialization plays a crucial role in training stability and performance. 
Transformer-based LLMs widely adopt small initialization\cite{nguyen2019transformers}, which initialze all the weight with a normal distribution of standard deviation  $\sqrt{\frac{2}{5d}}$, where $ d $ is the hidden dimension.
It's also common practice to scale the weights of residual layers at initialization by a factor of $ 1 / \sqrt{L}$ \cite{radford2019language}, where $ L $ is the number of
 layers.

Our findings suggest that scaling initialization by both model depth and width, using $ \sqrt{\frac{1}{2dL}}$, leads to faster loss convergence and improved performance on downstream tasks. We call this initialization method TinyInit.
We hypothesize that TinyInit achieves more consistent parameter scales across the model, which may facilitate optimization and convergence. 

Research \cite{takase2024spikemore} indicates that embedding layers require different initialization strategies compared to other layers.
Specifically, maintaining the standard deviation of embedding weights close to 1 may enhance training stability. 
Our experimental results indicate that initializing with a standard deviation of 0.5 achieves good model performance.

\subsection{Tokenizer}\label{section:tokenizer}
The design of the tokenizer significantly impacts model performance. An optimal vocabulary balances domain coverage (handling diverse tasks such as text, math, and code) with efficiency (encoding data with fewer tokens). Common methods use Byte-Pair Encoding (BPE)~\cite{shibata1999byte} and SentencePiece~\cite{kudo-richardson-2018-sentencepiece} build vocabularies by directly computing word frequencies across the entire training dataset. However, this approach suffers from domain imbalance, as common domains such as general text dominate the vocabulary, while specialized domains such as math and code remain underrepresented due to their limited data volume.

\modelname~adopts a domain-aware vocabulary strategy. We perform independent frequency analyses across multiple domains including general Chinese, general English, code, and mathematics, generating distinct domain-specific vocabularies. These vocabularies are then merged and de-duplicated to form a unified vocabulary of 153,376 unique tokens, maintaining balanced representation across domains while preserving overall compression efficiency. Table~\ref{tab:vocab_distribution} summarizes the detailed token distribution across different domains.
\begin{table}[htbp]
\centering
\caption{Token distribution in the unified vocabulary of \modelname.}
\label{tab:vocab_distribution}
\small
\begin{tabular}{lrr}
\toprule
\textbf{Domain} & \textbf{Number of Tokens} & \textbf{Percentage (\%)} \\
\midrule
English & 68,017 & 44.35 \\
Chinese & 41,053 & 26.77 \\
Other & 30,573 & 19.93 \\
Latin-based languages & 4,507 & 2.94 \\
Arabic & 2,755 & 1.80 \\
Korean & 2,733 & 1.78 \\
Mathematics & 2,139 & 1.39 \\
Japanese & 1,599 & 1.04 \\

\midrule
\textbf{Total} & \textbf{153,376} & \textbf{100.00} \\
\bottomrule
\end{tabular}
\end{table}

\section{Model Training}
In this section, we present our training pipeline, which is similar to training state-of-the-art language models, e.g., DeepSeek-V3~\cite{deepseekv3} and Llama 3~\cite{dubey2024llama}. The training process consists of three main stages: pre-training, long context extension, and post-training. Each stage has specific training strategies and data construction methods to gradually enhance the model capabilities.

\subsection{Pre-training Stage}

We first introduce the data construction in the pre-training of \modelname, followed by the details of data verification. Then we elaborate the practical approach for the long context extension. The detailed pre-training hyper-parameters are finally presented.

\subsubsection{Data Construction}
The pre-training corpus of \modelname~contains high-quality and diverse 13.2T tokens produced by our tokenizer, as stated in Section~\ref{section:tokenizer}.
Table~\ref{tab:data_recipe} shows the pre-training process is structured into three sequential phases: the \textit{general} phase, the \textit{reasoning} phase, and the \textit{annealing} phase. 
These phases are designed to progressively develop general knowledge and linguistic capabilities, enhance reasoning skills, and further refine knowledge and behavior, respectively.
The amount of data used in each phase is 12T, including 7.4T and 4.6T data in two distinct subphases, 0.8T, and 0.4T tokens.

\begin{table}[h!]
    \centering
        \caption{Data recipe of \modelname~pre-training.}
        \small
        \begin{tabular}{l|c|c|c} 
        \toprule
        \textbf{Dataset} & \textbf{General} & \textbf{Reasoning} & \textbf{Annealing} \\ 
        \midrule
        General English & 54\% & 14\% & 21\% \\ 
        General Chinese & 13\% & 6\% & 20\% \\ 
        Multi-lingual & 8\% & 4\% & 3\% \\ 
        Instruction & 2\% & 11\% & 20\% \\ 
        Math & 6\% & 28\% & 18\% \\ 
        Code & 17\% & 37\% & 18\% \\ 
        \bottomrule
        \end{tabular}
    \label{tab:data_recipe}
\end{table}

In the initial general training phase, we utilize a corpus focused on developing broad linguistic capabilities and general knowledge. 
This stage primarily consists of English and Chinese data collected from a diverse range of sources, including web pages, books, encyclopedias, \textit{etc}.
Data from the multilingual and various industrial domains is also incorporated. Based on our data quality assessment in Section~\ref{sec:data_quality}, we perfer to use higher-quality data in the second sub-phrase than the first.

In the second reasoning phase, we increase the proportion of high-quality and diverse mathematical and coding data---raising it to over 60\% of the corpus to enhance the reasoning capabilities of \modelname. 
The coding data includes both pure code and mixed text-code samples. The math data also involves a lot of English and Chinese texts.
Moreover, LLM-generated synthetic data is widely incorporated to enrich the corpus.

The third annealing phrase is designed to help the model consolidate and effectively apply the knowledge and reasoning skills acquired in the previous stages. 
Therefore, we place greater emphasis on instruction data, which accounts for approximately 20\% of the corpus. 
We curate in-house question banks covering a wide range of topics and construct both short and long chain-of-thought (CoT) responses. 
These reasoning paths are carefully refined to ensure clarity and logical coherence.

Overall, the pre-training data for \modelname~is carefully designed to ensure high quality, diversity, and minimal redundancy. 
We assign quality and difficulty labels to the data and adopt a curriculum-based sampling strategy for the reasoning data across all three phases---progressing from simpler examples to more complex ones throughout the training cycle.

\subsubsection{Data Quality Assessment}
\label{sec:data_quality}

Data quality assessment plays a crucial role in enhancing the overall quality of the data. 
Training \modelname~employs both rule-based heuristics and model-based evaluation to enhance data quality.

For model-based quality assessment, we leverage the Pangu series as the base model. To better align quality evaluation with human value judgments, we fine-tune the model using a manually annotated dataset. The fine-tuned evaluator is then applied to a large-scale pre-training corpus exceeding 10T tokens. Data samples are scored across multiple dimensions, including cleanliness, fluency, educational value, and richness. These annotated scores are then used in a prioritized sampling strategy, where higher-quality samples are assigned higher sampling probabilities.

To validate the effectiveness of our data quality assessment, we conducted an ablation study using a proxy model with 2.6 billion parameters. Empirical results show that, to achieve comparable performance, the model trained on low-scoring data required 1.6× more tokens than the one trained on high-quality high-scoring data. Therefore, high data quality is important for improving training efficiency.

\subsubsection{Pre-training Parameters} \label{sec:hyper_params}
\modelname~is trained using AdamW optimizer~\cite{loshchilov2017decoupled} with a weight decay of 0.1 and epsilon is set to $1 \times 10^{-8}$. 
The momentum parameters are set to $\beta_1 = 0.9$ and $\beta_2 = 0.95$. The gradient clipping norm is set to 1.0.
To improve the training stability and overall performance, the pre-training of \modelname~is organized into the following phases:

\textbf{0T--7.4T tokens} The sequence length is set to 4K (RoPE base = $1 \times 10^4$). The batch size increase from 1,024 to 1,536 (at 1.2T) and 2,048 (at 1.9T). The increased batch size improves training efficiency and throughput. The learning rate follows a cosine decay from $1 \times 10^{-4}$ to $1 \times 10^{-5}$ with 4,000 warmup steps to ensure stable early training.

\textbf{7.4T--12.0T tokens} The sequence length remains at 4K with a batch size of 2,048. The learning rate is fixed at $1 \times 10^{-5}$ in this phase.

\textbf{12.0T--12.8T tokens} The sequence length increases to 8K (RoPE base = $1 \times 10^5$). The batch size is reduced to 1,536. The learning rate decays from $1 \times 10^{-5}$ to $7.5 \times 10^{-6}$ using cosine scheduling.

\subsection{Long Context Extension}
The ability of LLMs to understand long context inputs is critical in long-thinking process and practical applications.
In the final stages of pre-training, \modelname~is trained on long sequence data to support a maximum context length of 128K.
The training consists of two progressive phases: the first phase expands the context length to 32K, and the second phase further expands it to 128K.

Rotary Position Embedding (RoPE)~\cite{su2023roformerenhancedtransformerrotary} is the core module for supporting ultra-long input sequences. 
Existing open-source LLMs typically extend context length by either increasing the base frequency in RoPE~\cite{su2023roformerenhancedtransformerrotary,qwen-2.5-coder} or by adopting methods such as YaRN~\cite{peng2023yarnefficientcontextwindow,dubey2024llama,deepseekv3}.
Our findings show that both methods perform similarly well if the hyper-parameters are correctly chosen, and we adopt the increased base frequency method in \modelname.
To determine the base frequency in RoPE for long-context extension, we evaluate the offline performance of ``Needle In A Haystack'' (NIAH) with different base frequencies at the target sequence length, and select the one with the best result.
This ensures a relatively low initial loss in long-context training.
In practice, the selected base frequency for 32K is $1.6\times 10^6$, and for 128K is $2.56\times 10^{7}$.
Detailed hyper-parameters of \modelname~long context training are summarized below:

\textbf{8K to 32K phase} The sequence length is expanded to 32K (RoPE base = $1.6\times 10^6$). The batch size is 384 with a learning rate of $7.5\times 10^{-6}$, matching the final learning rate from the previous post-training stage.

\textbf{32K to 128K phase} The sequence length is further expanded to 128K (RoPE base = $2.56\times 10^{7}$). The batch size is reduced to 96. The learning rate remains $7.5\times 10^{-6}$.

\subsection{Post-training Alignment} 
In the post-training stage, \modelname~is aligned with human preferences through Supervised Fine-Tuning (SFT) and Reinforcement Learning (RL). This stage focuses on constructing high-quality, diverse instruction data and designing scalable, efficient training strategies.

\subsubsection{Post-training Data}
In constructing post-training data, we emphasize the data quality, diversity, and complexity. 
The data pool is curated from a wide range of domains and task types, including general question answering, AI-generated content (AIGC), text classification and analysis, programming, mathematics, logical reasoning, and tool usage.  
These tasks cover application areas such as finance, healthcare, and public services. 
Data sources span open-source instruction datasets, real-world industrial queries, and synthetic problems derived from the pre-training corpus. 

To promote data diversity, data samples are selected along two orthogonal dimensions, guided by the entropy law~\cite{yin2024entropy}: domain and task type. 
Hierarchical tagging models with varying levels of granularity are used to support balanced data sampling. 
Data quality is managed through a combination of rule-based validation and model-based validation, which helps eliminate low-quality or ambiguous samples.

To better stimulate the reasoning capabilities of \modelname, a large portion of the post-training data, approximately six-sevenths, consists of reasoning tasks such as mathematics, coding, and logic. 
The post-training data covers a range of complexities, with a focus on moderately to highly challenging tasks.

\subsubsection{Post-training Strategy}
In the post-training stage, \modelname~was first trained with SFT to establish preliminary instruction-following capabilities.
Following SFT, we apply RL with outcome-based reward signals to further enhance reasoning, alignment, and instruction-following abilities of \modelname.

We implement a latency-tolerant reinforcement learning framework optimized for the Ascend infrastructure, which will be detailed in a future report.
The framework enables efficient large-scale policy optimization on Ascend.
To guide the RL process, we implement a hybrid reward system that provides task-specific feedback for mathematics, coding, and general problem-solving. 
This hybrid reward system combines deterministic reward signals and model-based evaluations to facilitate stable and efficient policy optimization.

\section{Training System}
Training our \modelname~with 135B parameters on 13.2 trillion tokens necessitates the need to ensure training stability and efficiency in large-scale computing cluster.
In this section, we elaborate the details of our training system from two important perspectives: parallelization strategies and system-level optimization techniques, in Section~\ref{sec: parallelism} and Section~\ref{sec: sys-optimization}. 
Overall, we achieve over 52\% Model FLOPs Utilization (MFU) when training \modelname{} on 8,192 Ascend NPUs.

\subsection{Computing Setup}
A computing cluster with 8,192 Ascend Neural Processing Units (NPUs)~\cite{910b2, 910b2-technical} is deployed to train \modelname.
Each node in the cluster houses 8 NPUs, interconnected via Huawei Cache Coherence System (HCCS) using a full-mesh topology, and each device is equipped with 64GB Memory.
Inter-node communication is facilitated through RDMA over Converged Ethernet (RoCE) fabric, leveraging 200 Gbps interconnects for communication between NPUs across different nodes.

\subsection{Parallelism Strategies for Model Scaling}  \label{sec: parallelism}
\begin{figure}[htbp]
    \centering
    \includegraphics[width=0.8\textwidth]{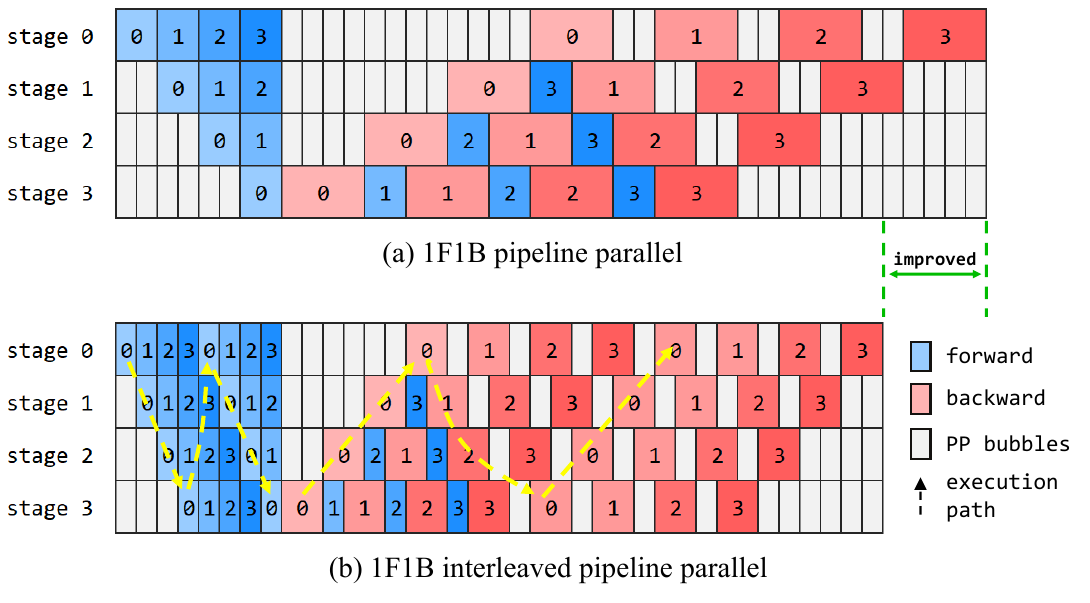}
    \caption{Pipeline parallelism and the interleaved pipeline-parallel scheduling.}
    \label{fig:interleaved-pp}
\end{figure}

In order to scale model training\footnote{The training of \modelname~is supported by MindSpeed \cite{mindspeed} and Megatron \cite{megatron-github, megatron-lm} framework, which provides comprehensive parallel strategies and system optimization methods.}, we leverage a combination of different parallelism strategies to distributes the model across multiple NPUs, including Data Parallelism (DP) \cite{data_parallel}, Tensor Parallelism (TP) \cite{megatron-lm}, Sequence Parallelism (SP) \cite{megatron-sp}, and Pipeline Parallelism (PP) \cite{gpipe, pipedream}.
For \modelname, 128-way DP with ZERO \cite{zero} is performed to reduce the memory cost of model parameters and the associated optimizer states. 
8-way TP is applied to leverage the high intra-node bandwidth for efficient activation transfer, while 8-way PP is adopted to utilize inter-node connections, since it only requires transmitting activations at the partition boundaries.
However, as mentioned in existing studies \cite{megascale, gpipe, pipedream, zero_pp}, pipeline parallelism encounters severe PP bubbles when the training cluster scales up, primarily due to batch size constraints \cite{megascale}.
For one-forward-one-backward (1F1B) PP scheduling, the bubble ratio is defined as $\frac{p-1}{p-1+n}$, where $p$ represents the number of pipeline stages and $n$ denotes the number of micro batches for every DP. The ratio represents the idle time of accelerators, as shown in Figure \ref{fig:interleaved-pp}.
A large-scale training cluster increases the number of DPs, which in turn reduces the number of micro batches assigned to each DP due to batch size constraints, leading to a significant increase in the bubble ratio.
Therefore, minimizing bubble ratio is crucial for improving system efficiency.
Under such circumstances, we employ interleaved pipeline-parallel scheduling with 6-way virtual PP stages on each device~\cite{interleaved-pp} and manage to reduce it from 30.45\% to 6.8\%. 
Through careful tuning of load balancing across PP and VPP stages, we are able to achieve approximately 43\% MFU on an 8,192 NPU cluster as a baseline.

\subsection{System Optimization}\label{sec: sys-optimization}

Based on the optimizations outlined in Section \ref{sec: parallelism} that achieved 43\% MFU, additional system-level enhancements are implemented to push training efficiency to new heights. 
Through a combination of kernel fusions, context parallelism via subsequence partitioning, data caching and sharing mechanisms, and other refinements, \modelname~benefits from a significant improvement in training efficiency.
These comprehensive optimizations enable the system to achieve over 52\% MFU, representing a 9\% relative improvement compared to the baseline configuration mentioned in Section \ref{sec: parallelism}.

\begin{figure}[htbp]
    \centering
    \includegraphics[width=0.7\textwidth]{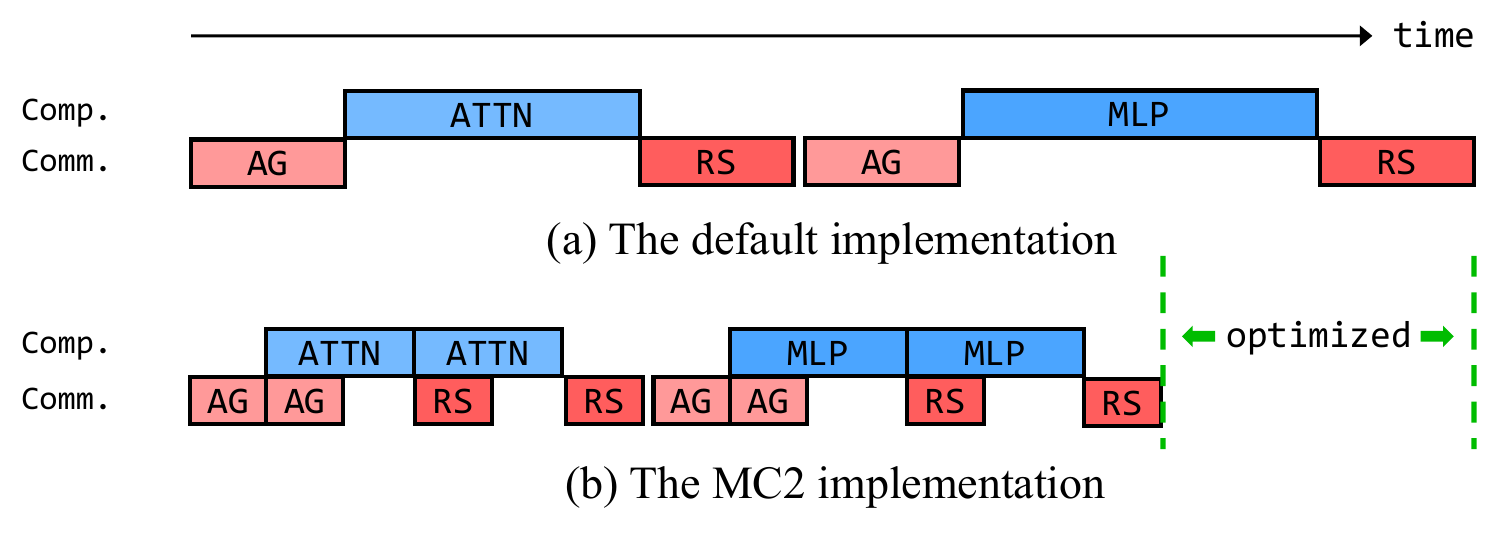}
    \caption{A Comparison of the default transformer computation and the MC2 method. Note that in actual training, communication and computation tasks are fused into a single kernel in MC2.} 
    \label{fig:mc2}
\end{figure}

\subsubsection{Kernel Fusion}
Kernel fusion is widely adopted in LLM training to enhance efficiency. It combines multiple operations into a single kernel, reducing the number of data accesses to global memory \cite{flash_attention2}. 
During the training phase of \modelname, key operators are fused, resulting in significant improvements in hardware utilization and overall training efficiency.

\textbf{MC2 - Merged Compute and Communication} 
Tensor parallelism, when combined with sequence parallelism, introduces All-Gather (AG) and Reduce-Scatter (RS) communication operations for exchanging input and output activations across distributed devices.
This approach exhibits a direct dependency between matrix multiplication (MatMul) and AG/RS communications, which fundamentally constrains the overlapping of TP communication with computational workflows.
The MC2 is implemented \cite{ascend_mc2, mc2} to tackle this challenge by fusing MatMul computations with communication operations.
It decomposes large computation and communication tasks into fine-grained subtasks and employs pipelined execution to maximize overlap between communication and computation. 
Thus, MC2 significantly reduces communication latency and improves hardware utilization (Figure \ref{fig:mc2}).

\textbf{NPU Fusion Attention}
Training LLMs with long sequence length suffers from quadratic memory and computational requirements in self-attention mechanisms as sequence length grows. 
To address these challenges, Flash Attention (FA) has emerged as a standard technique in LLM training owing to its superior performance \cite{flash_attention, flash_attention2}.
\modelname~ leverages a self-attention fusion operator, called NPU Fusion Attention (NFA)\cite{npu_fa}, which is specifically optimized for Ascend NPUs, offering system-level improvements across a wide range of self-attention computation scenarios.
\begin{figure}[htbp]
    \centering
    \includegraphics[width=0.4\textwidth]{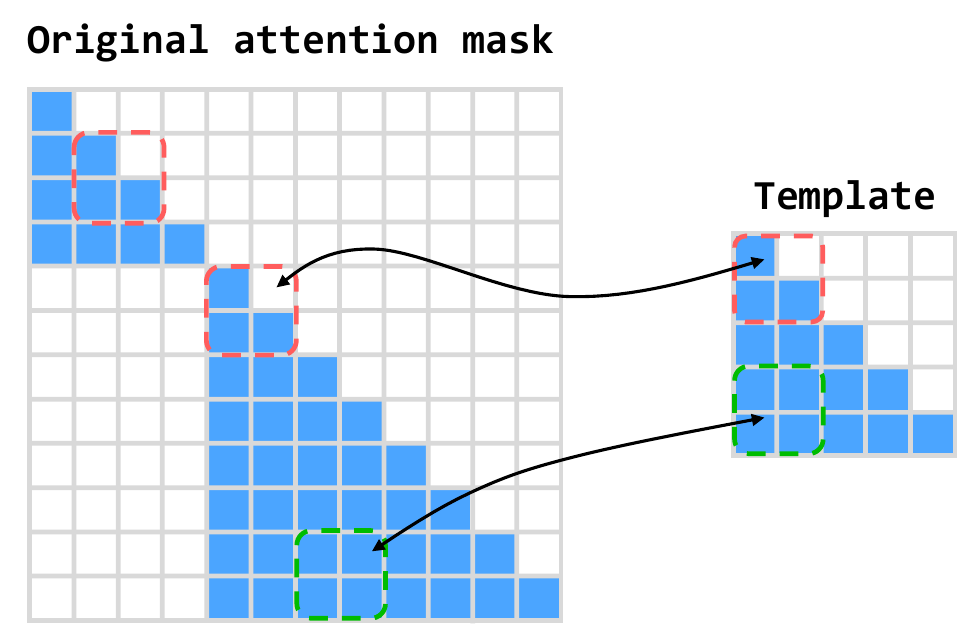}
    \caption{Examples of attention mask compression for the NFA operator.}
    \label{fig:attn_mask_compression}
\end{figure}

It is worth mentioning that \modelname~uses a reset attention mask strategy to prevent self-attention between different documents within a sequence.
This requires calculating the corresponding attention mask for every sequence, leading to significant memory and computational overhead.
To mitigate the time and memory requirements of generating attention masks, the NFA operator employs a mask compression optimization.
As shown in Figure \ref{fig:attn_mask_compression}, NFA utilizes a $2048 \times 2048$ causal mask as a template to construct the computational mask within the fusion attention operator.
For every iteration, \modelname~ retrieves the actual sequence length based on the position of the end-of-document (eod) token, which is then provided as input to the NFA operator to accelerate the computation of self-attention.
The detailed usage of NFA is provided in the Ascend documentation~\cite{npu_fa}.

\textbf{Other Kernel Fusions for Efficiency} 
In addition to MC2 and NPU-optimized fused attention, we also integrate a series of kernel fusion optimizations within key components such as RMSNorm \cite{zhang2019rootmeansquarelayer}, SwiGLU \cite{shazeer2020glu}, and rotary positional embeddings (RoPE) \cite{su2023roformerenhancedtransformerrotary}, as well as critical processes including gradient accumulation and PP send/receive communications.
These fusion operators are designed to reduce kernel launch and memory access overheads, while maintaining high numerical precision and enhancing overall training performance.

\begin{figure}[htbp]
    \centering
    \includegraphics[width=\textwidth]{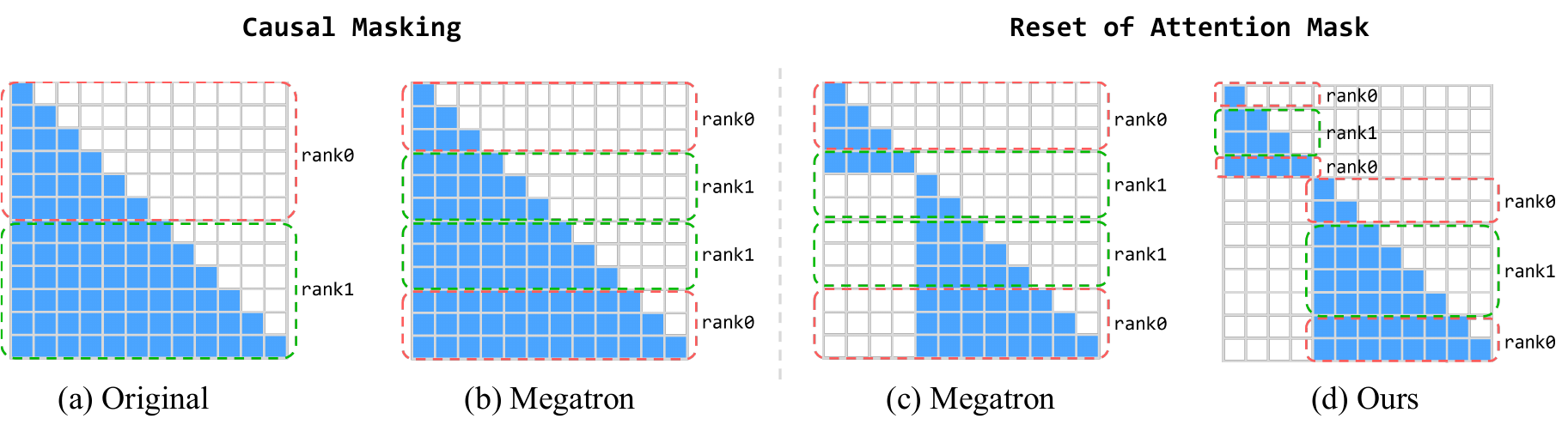}
    \caption{Examples of the mechanism of sub-sequence partitioning for context parallelism.}
    \label{fig:sub_seq_partition}
\end{figure}

\subsubsection{Optimization for Long Context Training}
Scaling long-context capabilities is becoming increasingly important for applications such as long document summarization and conversational AI.
However, training on long sequences presents several challenges in terms of both time and memory complexity.
To improve the efficiency of long-context training, we propose two key strategies, as outlined below.

\textbf{Sub-Sequence Partitioning for Context Parallelism} 
Context parallelism (CP) is an crucial approach for the training of very long sequences, that divides the input sequence into segments to reduce memory consumption \cite{ringattention, ulysses}.
Yet, with causal masking, simply splitting the sequence into $CP$ chunks results in a severely imbalanced workload for Ring Self-Attention (RSA) \cite{ringattention} (as shown in Figure \ref{fig:sub_seq_partition}(a)).
Megatron-LM addresses this issue by splitting the sequence into $2\times CP$ chunks, where each rank receives chunks from both the top and bottom, thus balancing the workload within a CP group (Figure \ref{fig:sub_seq_partition}(b)) \cite{megatron-github}.
However, this method still results in an imbalanced workload when the attention mask is reset (Figure \ref{fig:sub_seq_partition}(c)).
Therefore, in training with 128k-long contexts, we propose a load-balanced partitioning strategy for CP training, where each rank is responsible for computing two chunks within each subsequence (Figure \ref{fig:sub_seq_partition}(d)).

\textbf{Fast Mask Generation and Data Reuse}
When scaling the training sequence of \modelname~up to 128k, the generation of the attention mask or the calculation of the actual sequence length still incurs a non-negligible performance overhead.
Additionally, in the training scenario with reset attention masks, each VPP stage is required to retrieve the corresponding mask or actual sequence length in every iteration, resulting in redundant computations and increased overhead.
We optimize these problems by (1) using efficient NPU operators to compute the attention mask, instead of constructing it on the CPU, thus accelerating mask generation and eliminating the need for data transfer between the CPU and NPU, and (2) enabling cross-VPP stage mask sharing, where attention masks are generated by the first stage (VPP0) and shared across different VPP stages on the same rank, thereby avoiding redundant mask computations and memory cost.
\section{Results}
In this section, we discuss the evaluation results of \modelname, including pre-training performance and post-training outcomes. 
In addition, we provide comprehensive ablation studies that exam the model architecture and further discuss the observations of training \modelname.

\subsection{Pre-Training Training Loss Curve}
\begin{figure}[htbp]
    \centering
    \includegraphics[width=0.7\textwidth]{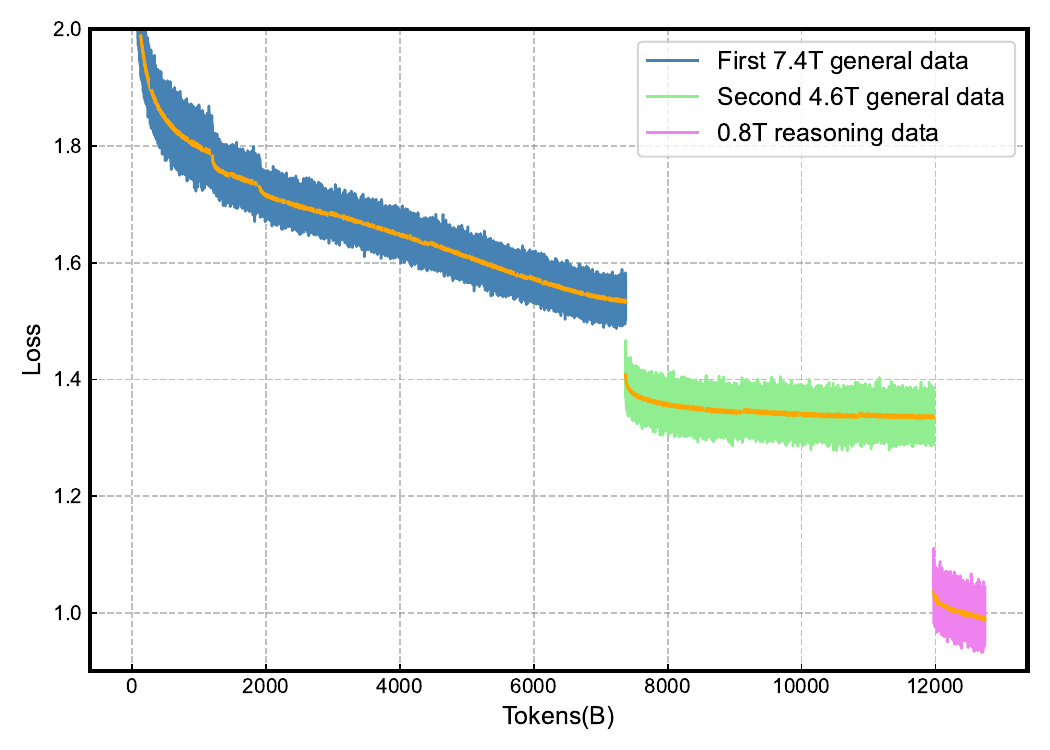}
    \caption{The training loss curve of \modelname~during the pre-training stage.}
    \label{fig:135b_loss}
\end{figure}

Figure~\ref{fig:135b_loss} shows the training loss curve of \modelname~during the entire pre-training. 
Each segment in the loss curve corresponds to one training stage, as described in Section~\ref{sec:hyper_params}. The loss curves demonstrate consistent descending trends across all training stages. For the second interval, although the descent rate moderated due to a constant learning rate, the performance metrics continued to show steady improvement throughout this interval.

\paragraph{Zero loss spike} As shown in Figure~\ref{fig:135b_loss}, \textbf{no loss spikes} occur throughout the entire pre-training process.
While such spikes are common in LLM training \cite{takase2024spikemore}, the absence of them here underscores the importance of our depth-scaled sandwich norm and TinyInit in ensuring stable training.
The negative effect of loss spike to the model performance will be further elaborated in Section~\ref{sec:sandwich-ablation}.

\subsection{Pre-Training Stage}
\paragraph{Benchmarks} We evaluate \modelname{} base model across multiple domains using open-source benchmarks, including language understanding, question answering, code generation, and math problem solving. The evaluation mainly uses English and Chinese test sets, with some additional multilingual benchmarks for broader coverage.

\begin{itemize}[leftmargin=*]

\item Language understanding: We employ \textit{Hellaswag} \cite{Zellers2019HellaSwagCA} and \textit{Winogrande} for contextual reasoning tasks, \textit{DROP} \cite{Dua2019DROPAR}, \textit{RACE} \cite{Lai2017RACELR}, and \textit{ARC} \cite{Clark2018ThinkYH} series for comprehensive reading comprehension evaluation, along with \textit{PIQA} \cite{Bisk2019PIQARA}, \textit{Natural Questions} \cite{Kwiatkowski2019NaturalQA} and \textit{TriviaQA} \cite{Joshi2017TriviaQAAL} to assess knowledge retrieval.

\item Question answering: The assessment includes \textit{C-Eval} \cite{Huang2023CEvalAM} for Chinese knowledge, \textit{MMLU} \cite{Hendrycks2020MeasuringMM} and its advanced variant \textit{MMLU-Pro} \cite{Wang2024MMLUProAM} for English domain knowledge, supplemented by \textit{BigBenchHard} \cite{Suzgun2022ChallengingBT} to evaluate creative problem-solving

\item Code generation and understanding: We utilize \textit{HumanEval} \cite{Chen2021EvaluatingLL} and \textit{MBPP} \cite{Austin2021ProgramSW} for standard code generation tasks, while \textit{CruxEval} \cite{Gu2024CRUXEvalAB} for code understanding and reasoning.

\item Mathematical Reasoning : We measure skills with \textit{CMath} \cite{Wei2023CMATHCY} and \textit{GSM8K} \cite{Cobbe2021TrainingVT} for fundamental arithmetic and simple problems, \textit{MATH} \cite{Hendrycks2021MeasuringMP} for advanced mathematical reasoning, and \textit{MGSM} \cite{Shi2022LanguageMA} for multilingual math problem solving.

\end{itemize}

\paragraph{Baselines \& Comparison Settings} We compare \modelname~against several strong baselines covers both dense models (Qwen2.5-72B, Llama-405B) and MoE architectures (DeepSeek-V3). For base models, the majority of our evaluations employ few-shot inputs, with a minority using zero-shot prompts. We evaluate most benchmarks with gold answers through exact matching, while employing execution-based verification for code generation tasks.

\paragraph{Evaluation Results}
In Table~\ref{tab:main}, we compare the pre-trained base model of \modelname~with other leading models. Overall, \modelname~achieves state-of-the-art performance on most general English benchmarks and all Chinese benchmarks.
While it trails DeepSeek V3 on code and math-related tasks, it performs competitively on these domains.

A closer examination reveals that \modelname~excels on Chinese benchmarks, surpassing both Qwen 2.5 72B and DeepSeek V3, the current best-performing Chinese model. 
In addition, when compared to Llama 3.1 405B, \modelname~achieves better scores on most of the challenging benchmarks, while utilizing only about 29\% of the training FLOPs required by Llama 405B.
These results suggest the effectiveness of our model architecture and the high quality of our training data.

\begin{table}[!h]
    \centering
    \footnotesize
    \setlength{\tabcolsep}{4.5pt}
    \caption{Comparison of \modelname~and other representative models across a diverse set of benchmarks for evaluating language, coding and mathematical skills. Bold values represent the best results in each line, and underlined values represent \modelname~is the best among dense models.}
    \small
    \begin{tabular}{@{}c l c | c c c | c@{}}
    \toprule
    & \multirow{2}{*}{\centering \textbf{Benchmark {\tiny (Metric)}}} & \multirow{2}{*}{\textbf{\# Shots}} & \textbf{Qwen2.5} & \textbf{Llama-3.1} & \textbf{DeepSeek} & \textbf{\modelname} \\
    & & & \textbf{72B Base} & \textbf{405B Base} & \textbf{V3 Base} & \textbf{Base} \\
    \midrule
    & Architecture & - & Dense & Dense & MoE & Dense \\
    & \# Activated Params & - & 72B & 405B & 37B & 135B \\
    & \# Total Params & - & 72B & 405B & 671B & 135B \\
    \midrule
    \multirow{16}{*}{English} & BBH {\tiny (EM)} & 3-shot & 79.8 & 82.9 & \textbf{87.5} & 79.1 \\
    & MMLU {\tiny (EM)} & 5-shot & 85.0 & 84.4 & \textbf{87.1} & \underline{85.4}  \\
    & MMLU-Pro {\tiny (EM)} & 5-shot & 58.3 & 52.8 & \textbf{64.4} & \underline{63.1}  \\
    & DROP {\tiny (F1)} & 3-shot & 80.6 & 86.0 & \textbf{89.0} & 61.0  \\
    & ARC-Easy {\tiny (EM)} & 25-shot & 98.4 & 98.4 & 98.9 & \textbf{100.0} \\
    & ARC-Challenge {\tiny (EM)} & 25-shot & 94.5 & 95.3 & 95.3 & \textbf{97.0}  \\
    & HellaSwag {\tiny (EM)} & 10-shot & 84.8 & 89.2 & 88.9 & \textbf{99.0}  \\
    & PIQA {\tiny (EM)} & 0-shot & 82.6 & 85.9 & 84.7 & \textbf{98.0} \\
    & WinoGrande {\tiny (EM)} & 5-shot & 82.3 & 85.2 & 84.9 & \textbf{91.0}  \\
    & RACE-Middle {\tiny (EM)} & 5-shot & 68.1 & 74.2 & 67.1 & \textbf{97.0} \\
    & RACE-High {\tiny (EM)} & 5-shot & 50.3 & 56.8 & 51.3 & \textbf{97.0} \\
    & TriviaQA {\tiny (EM)} & 5-shot& 71.9 & 82.7 & 82.9 & \textbf{90.5} \\
    & NaturalQuestions {\tiny (EM)} & 5-shot & 33.2 & 41.5 & 40.0 & \textbf{52.7} \\
    & AGIEval {\tiny (EM)} & 0-shot & 75.8 & 60.6 & 79.6 & \textbf{80.4} \\
    \midrule
    \multirow{4}{*}{Code} & HumanEval {\tiny (Pass@1)} & 0-shot & 53.0 & 54.9 & 65.2 & \textbf{81.1} \\
    & MBPP {\tiny (Pass@1)} & 3-shot & 72.6 & 68.4 & \textbf{75.4} & 72  \\
    & CRUXEval-I {\tiny (EM)} & 2-shot & 59.1 & 58.5 & \textbf{67.3} & \underline{61.8}  \\
    & CRUXEval-O {\tiny (EM)} & 2-shot & 59.9 & 59.9 & \textbf{69.8} & \underline{61.5}  \\
    \midrule
    \multirow{3}{*}{Math} 
    & GSM8K {\tiny (EM)} & 8-shot & 88.3 & 83.5 & \textbf{89.3} & \textbf{89.3}  \\
    & MATH {\tiny (EM)} & 4-shot & 54.4 & 49.0 & 61.6 & \textbf{62.5}  \\
    & MGSM {\tiny (EM)} & 8-shot & 76.2 & 69.9 & \textbf{79.8} & 75.1  \\
    & CMath {\tiny (EM)} & 3-shot & 84.5 & 77.3 & \textbf{90.7} & 78.2  \\
    \midrule
    \multirow{7}{*}{Chinese} & CLUEWSC {\tiny (EM)} & 5-shot & 82.5 & 83.0 & 82.7 & \textbf{95.0} \\
    & C-Eval {\tiny (EM)} & 5-shot & 89.2 & 72.5 & 90.1 & \textbf{90.3}  \\
    & CMMLU {\tiny (EM)} & 5-shot & 89.5 & 73.7 & 88.8 & \textbf{91.7} \\
    & CMRC {\tiny (EM)} & 1-shot & 75.8 & 76.0 & 76.3 & \textbf{86.0}  \\
    & C3 {\tiny (EM)} & 0-shot & 76.7 & 79.7 & 78.6 & \textbf{99.0} \\
    & CCPM {\tiny (EM)} & 0-shot & 88.5 & 78.6 & 92.0 & \textbf{93.0}  \\
    \bottomrule
    \end{tabular}
    \label{tab:main}
\end{table}

\subsection{Post-Training and Reasoning Capability}
\paragraph{Benchmarks} We conduct a comprehensive evaluation of the \modelname's capabilities over reasoning and non-reasoning tasks:

\begin{itemize}[leftmargin=*]
\item Sophisticated reasoning tasks encompass three specialized subcategories: mathematical competence measured by \textit{AIME 2024}~\cite{MAA} and \textit{MATH-500}, Coding competition benchmarks \textit{LiveCodeBench}~\cite{jain2024livecodebench} and scientific reasoning task \textit{GPQA Diamond}~\cite{rein2024gpqa};
\item General language comprehension and reasoning capabilities, represented by \textit{MMLU-Pro}~\cite{gema2024we}, \textit{Arena Hard}~\cite{arenahard2024}.
\end{itemize}

\paragraph{Baselines \& Comparison Settings} We compare \modelname~against strong baselines including GPT-4o-0513, reasoning models DeepSeek-R1, Hunyuan-T1 and large dense models, Qwen2.5-72B-Instruct and Mistral-Large 2. 
We use Pass@1  averaged over multiple independent runs as the evaluation metric to assess the performance.

\paragraph{Evaluation Results}
In Table~\ref{tab:StandardBenchmarks}, we compare the evaluation results of \modelname~with other baseline models. \modelname~achieves state-of-the-art performance on the reasoning benchmarks including AIME 2024, MATH-500, GPQA and LiveCodeBench, while maintaining strong capabilities in general language comprehension tasks.

When compared to dense LLMs (Qwen and Mistral-Large 2), \modelname~shows particularly significant advantages in reasoning tasks. This superior performance stems from the 0.8T reasoning-focused data used in pre-training (Section~\ref{sec:hyper_params}). The reasoning-enhanced base model substantially benefits subsequent post-training phases.

\begin{table}[!h]
    \centering
    \caption{Comparison of \modelname~models and other representative models across benchmarks. $\dagger$ indicates results from Artificial Analysis \cite{ai_analysis}.}
    \resizebox{\linewidth}{!}{
    \small
    \begin{tabular}{@{}l *{6}{c} @{}}
    
    \toprule
    \multirow{2}{*}{\centering\textbf{Model}} & \multirow{2}{*}{\textbf{AIME 2024}} & \multirow{2}{*}{\textbf{MATH-500}} & \textbf{GPQA} & \textbf{LiveCode} & \multirow{2}{*}{\textbf{ArenaHard}}& \multirow{2}{*}{\textbf{MMLU-pro}} \\
    &  &  & \textbf{Diamond} & \textbf{Bench} &  \\
     \midrule
    \textbf{GPT-4o-0513} & 9.3 & 74.6 & 49.9 & 32.9 & 80.4 & 72.6 \\
    \textbf{Qwen2.5-72B} & 16.0 & 83.1 & 49 & 27.6 & 81.2 & 72.0 \\
    \textbf{Mistral-Large 2}$^\dagger$ & 11.0 & 73.6 & 48.6 & 29.3 & - & 69.7 \\
    \textbf{Hunyuan-T1} & 79.8 & 96.2 & 69.3 & 64.9 & 91.9 & \textbf{87.2} \\
    \textbf{DeepSeek-R1} & 79.8 & 97.3 & 71.5 &  65.9 & \textbf{92.3} &  84.0\\
    \midrule
    \textbf{\modelname} & \textbf{80.8} & \textbf{97.4} & \textbf{74.2} & \textbf{66.5} & 91.5 & 84.4 \\
    \bottomrule
    \end{tabular}
    }
    \label{tab:StandardBenchmarks}
\end{table}

\subsection{Ablation Studies}  
This section presents additional ablation studies of the model architecture and analyzes key training behaviors to facilitate a deeper understanding and discussion of dense LLM training.

\subsubsection{Depth-scaled Sandwich-norm} \label{sec:sandwich-ablation}
We conducted experiments to validate the effectiveness of depth-scaled sandwich norm compared to pre-norm architectures. Using a dense Transformer model with 13 billion parameters trained on 300 billion tokens with identical hyperparameters for both configurations, we observe significant improvements.

\begin{figure}[htbp]
  \centering
  \subfloat[Loss]{
    \includegraphics[width=0.47\textwidth]{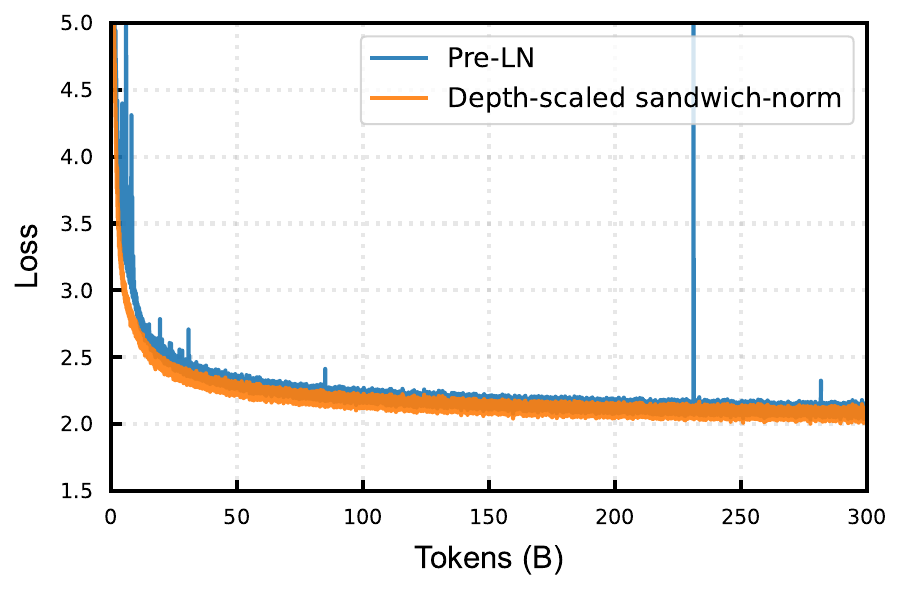}
  }
  \hfill
  \subfloat[Gradient norm]{
    \includegraphics[width=0.48\textwidth]{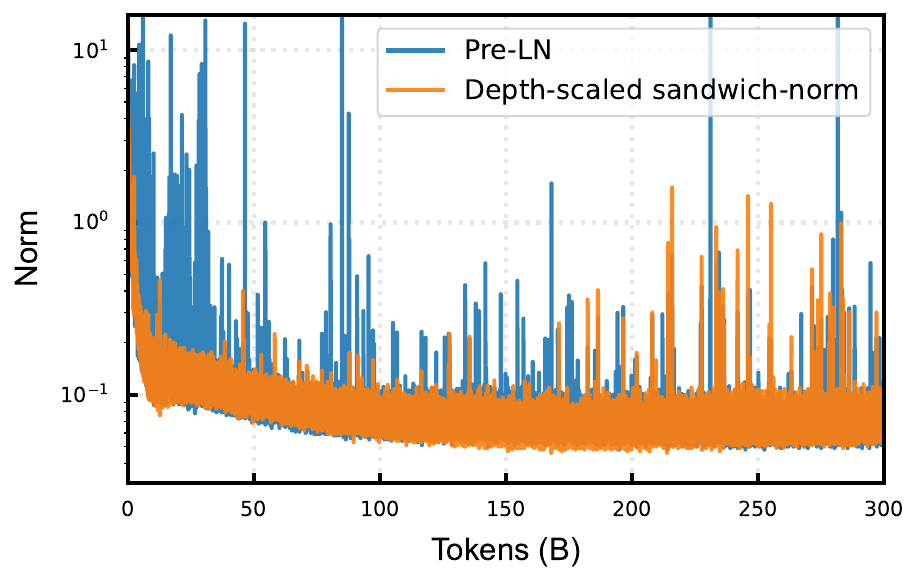}
  }
  \caption{Pre-training loss and gradient norm for a 13B model using Pre-LN and Depth-Scaled Sandwich-Norm (DSSN). The curves with Pre-LN has significant spikes, which harm the trained model, while the curves of DSSN are much smoother.  }
  \label{fig:sandwich_compare}
\end{figure}

Figure~\ref{fig:sandwich_compare} shows the depth-scaled sandwich-norm architecture stabilizes gradient norms and effectively eliminates loss spikes, leading to faster training convergence.
We evaluated performance on two composite benchmarks: EN basic, consisting of multiple English benchmarks, and ZH basic, representing Chinese benchmarks. Additional evaluation using LAMBADA \cite{Paperno2016TheLD} (English) and WPLC \cite{Ge2021ChineseWA} (Chinese) next-token prediction tasks confirmed the advantage of applying depth-scaled sandwich-norm. 
The results clearly suggest that preventing loss spikes during pre-training is crucial for optimal model performance.

\begin{table}[h]
	\centering
    \small
	\caption{Performance comparison between Pre-LN and Depth-scaled Sandwich-Norm.}
    \label{tab:sandwich_compare}
	\setlength{\tabcolsep}{8.0pt}
	\begin{tabular}{cccccc}
		\toprule
		\textbf{Model} & \textbf{Tokens (B)} & \textbf{EN basic} & \textbf{ZH basic} & \textbf{LAMBADA} & \textbf{WPLC} \\
		\midrule
		Pre-LN & 300 & 0.42 & 0.52 & 0.675 & 0.194 \\
		Depth-scaled sandwich-norm & 300 & 0.45 & 0.54 & 0.693 & 0.224 \\
		\bottomrule
	\end{tabular}
\end{table}

\begin{figure}[htbp]
    \centering
    \includegraphics[width=0.6\textwidth]{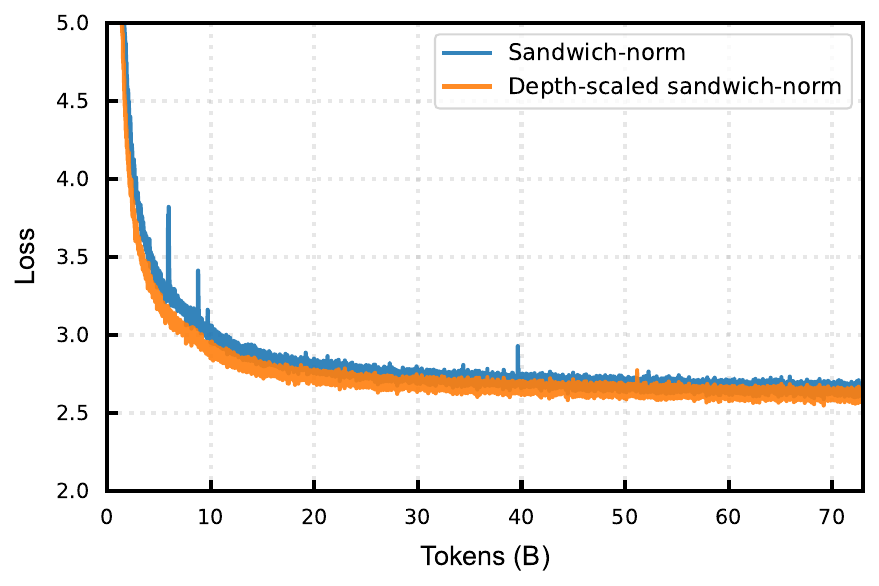}
    \caption{Pre-training loss for a 94-layer 1.6B model using original and depth-scaled sandwich-norm. The original sandwich-norm still suffers loss spikes during training.}
    \label{fig:1.6B_94l}
\end{figure}

To further ablate the effect of our depth-scaled factor in RMSNorm initialization, we compare with the plain sandwich-norm that does not have the $\sqrt{L}$ scaling factor in Eq.~\eqref{eq:sandwich}. 
Here, we use a proxy model containing 1.6 billion parameters and 94 layers, which has the same depth with \modelname. 
By using this proxy model, we examine the effectiveness of sandwich-norm on training very deep Transformers. 
In Figure~\ref{fig:1.6B_94l}, we can observe some loss spikes with the plain sandwich-norm, but our depth-scaled sandwich-norm can be trained smoothly, and attains lower loss.

\subsubsection{Tiny Initialization}
We conduct experiments to study the effectiveness of TinyInit proposed in Section~\ref{sec:tinyinit}.
After being trained on 102 billion tokens, \modelname{} initialized with TinyInit strategy, with standard deviation $\sqrt{\frac{1}{2dL}}$), performs significantly better than the baseline model that utilizes traditional initialization, whose standard deviations are $\sqrt{\frac{2}{5d}}$ and $\sqrt{\frac{2}{5dL}}$. The results are shown in Table~\ref{tab:tinyinit_exp}. BIG-bench (aug) is a test set developed internally through data augmentation of the original BIG-bench, designed to mitigate the impact of test set leakage.
\begin{table}[h]
	\centering
        \small
	\caption{Performance comparison of traditional initialization and TinyInit.} \label{tab:tinyinit_exp}
	\begin{tabular}{ccccccccc}
		\toprule
		\textbf{Model} & \textbf{Tokens (B)} & \textbf{EN basic} & \textbf{ZH basic} & \textbf{LAMBADA} & \textbf{WPLC} & \textbf{C-Eval} & \textbf{MMLU} & \textbf{BIG-bench (aug)} \\
		\midrule
		Baseline & 102 & 0.444 & 0.538 & 0.694 & 0.229 & 0.476 & 0.473 & 0.357 \\
		TinyInit & 102 & 0.456 & 0.537 & 0.727 & 0.257 & 0.524 & 0.502 & 0.384 \\
		\bottomrule
	\end{tabular}
\end{table}

\subsubsection{Layer Statistics of \modelname}

\begin{figure}[htbp]
    \centering
    \includegraphics[width=\textwidth]{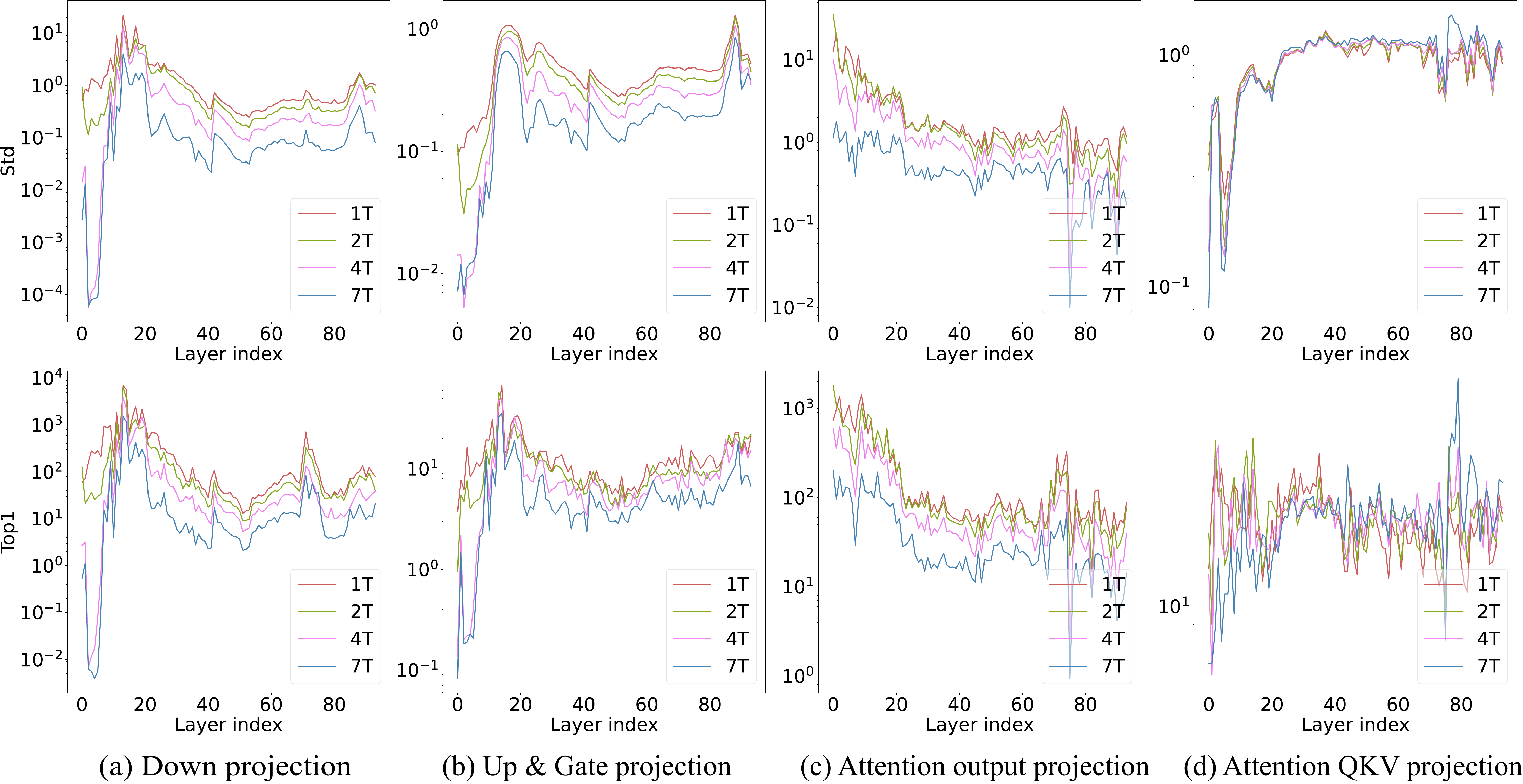}
    \caption{Activation of attention and FFN modules. Mean, standard deviation, and top-1 value of activations are included. Each line represents different training tokens from 1T, 2T, 4T to 7T. The "Std" row shows the stable activation scale across layers. The "Top 1" row shows the existence of the "super activations" in down projection and attention output projection, with magnitudes falling within a reasonable range and comparable to those observed in the LLaMA model~\cite{Yu2024TheSW}.}
    \label{fig:act_analysis}
\end{figure}

\paragraph{Stable activation scale}
Figure~\ref{fig:act_analysis} presents the activation patterns of attention and FFN modules across Transformer layers, showing the mean, standard deviation, and top-1 activation values. The activation distributions demonstrate stability, with standard deviations maintaining consistent scales throughout pre-training while preserving a clear layer-wise pattern.
Our analysis reveals the presence of ``super activations'', whose magnitude reaches $10^3$ magnitude in shallow layers, a phenomenon consistent with findings in the Llama model~\cite{Yu2024TheSW}.
Notably, Figure~\ref{fig:act_analysis} illustrates that these top-1 activation values progressively decrease with layer depth, indicating that their influence becomes relatively small on the final output.

\paragraph{Layer-wise patterns of depth-scaled sandwich norm.}

\begin{figure}[htbp]
    \centering
    \includegraphics[width=\textwidth]{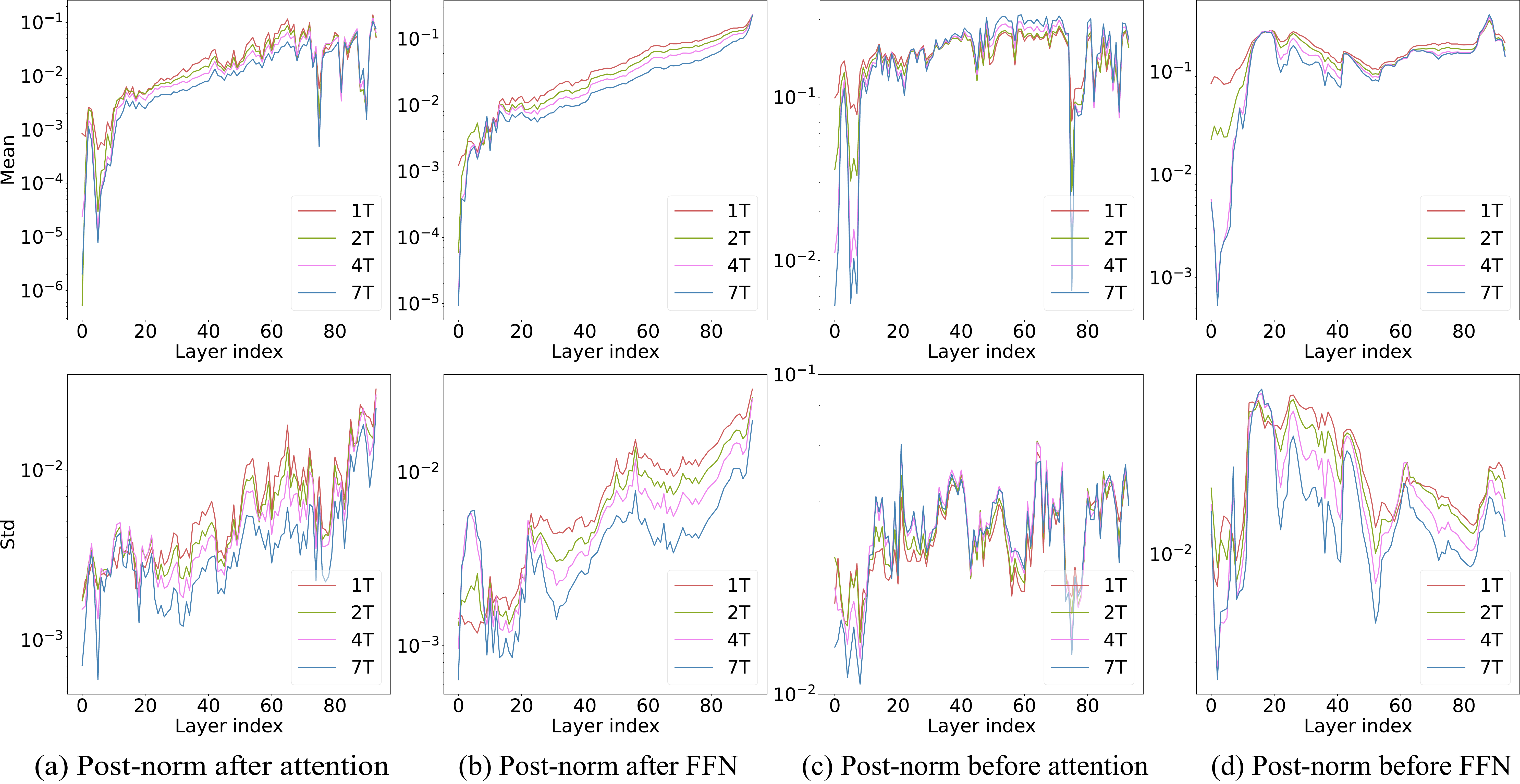}
    \caption{Distribution of sandwich-norm's $\gamma$ parameter. Mean and standard deviation are included. Each line represents different training tokens from 1T, 2T, 4T to 7T. There is a clear layer-wise pattern of the two post-norms: the mean and std value of $\gamma$ increase with depth. Larger post-norm $\gamma$ indicates deeper layers emphasize more on transformer outputs instead of residual connections.}
    \label{fig:gamma_analysis}
\end{figure}

Figure~\ref{fig:gamma_analysis} presents the distribution of scaling parameters $\gamma$ across all sandwich-norm layers, revealing several key observations:
All four LayerNorm $\gamma$ parameters exhibit decreasing mean/standard deviation during training, consistent with weight decay effects.
Post-norm $\gamma$ values show layer-dependent patterns:
The standard deviation of post-norm $\gamma$ increases substantially with layer depth.
Pre-norm $\gamma$ maintains relatively constant standard deviation across layers.
This pattern suggests an intriguing model behavior: shallow layers rely primarily on residual connections, while deeper layers progressively emphasize transformer layer outputs as the scaling factor $\gamma$ grows in magnitude.

\section{Conclusion}

We present \modelname{}, a dense language foundation model with 135 billion parameters trained on Ascend NPUs. To address challenges in training large-scale deep models, we propose depth-scaled sandwich-norm, enabling \modelname{} to achieve remarkable training stability without significant loss spikes. After being pre-trained on 13.2 trillion tokens and long context extension on 8,192 Ascend NPUs, our model further enhances its reasoning capabilities through Supervised Fine-Tuning and Reinforcement Learning. Extensive experiments lead to the observation that \modelname{} not only surpasses state-of-the-art dense LLMs like Llama 405B and Mistral Large 2 but also delivers competitive performance against larger sparse models such as DeepSeek-R1. These results highlight the efficacy of our architectural and systemic optimizations, paving the way for future advancements in scalable and efficient LLM training. In addition, our experience demonstrates that the Ascend NPUs are capable of training dense models with hundreds of billions of parameters.

\bibliographystyle{plain}
\bibliography{ref}  

\newpage
\appendix
\section{Contributions and Acknowledgments}

\noindent
\textbf{Core Contributors}
Yichun Yin, Wenyong Huang, Kaikai Song, Yehui Tang, Xueyu Wu, Wei Guo, Peng Guo, Yaoyuan Wang, Xiaojun Meng, Yasheng Wang, Dong Li, Can Chen, Dandan Tu, Yin Li, Fisher Yu, Ruiming Tang, Yunhe Wang

\textbf{Contributors}
Baojun Wang, Bin Wang, Bo Wang, Boxiao Liu, Changzheng Zhang, Duyu Tang, Fei Mi, Hui Jin, Jiansheng Wei, Jiarui Qin, Jinpeng Li, Jun Zhao, Liqun Deng, Lin Li, Minghui Xu, Naifu Zhang, Nianzu Zheng, Qiang Li, Rongju Ruan, Shengjun Cheng, Tianyu Guo, Wei He, Wei Li, Weiwen Liu, Wulong Liu, Xinyi Dai, Yonghan Dong, Yu Pan, Yue Li, Yufei Wang, Yujun Li, Yunsheng Ni, Zhe Liu, Zhenhe Zhang, Zhicheng Liu


\end{document}